# Semi-supervised Data Representation via Affinity Graph Learning

Weiya Ren[1]

[1] College of Information System and Management, National University of Defense Technology, Changsha, Hunan, P.R China, 410073
E-mail: weiyren.phd@gmail.com

*Abstract*—**We consider the general problem of utilizing both labeled and unlabeled data to improve data representation performance. A new semi-supervised learning framework is proposed by combing manifold regularization and data representation methods such as Non negative matrix factorization and sparse coding. We adopt unsupervised data representation methods as the learning machines because they do not depend on the labeled data, which can improve machine's generation ability as much as possible. The proposed framework forms the Laplacian regularizer through learning the affinity graph. We incorporate the new Laplacian regularizer into the unsupervised data representation to smooth the low dimensional representation of data and make use of label information. Experimental results on several real benchmark datasets indicate that our semi-supervised learning framework achieves encouraging results compared with state-of-art methods.**

*Index Terms*—**Semi-supervised learning, sparse coding, clustering, manifold regularization, nonnegative matrix factorization, metric learning.**

## I. INTRODUCTION

The task of semi-supervised learning (SSL) algorithms is to utilize both labeled and unlabeled data to improve learning ability. A variety of graph-based semi-supervised learning (GSSL) [1,5,8,9,10,13] have recently become popular due to their high accuracy and computational efficiency. Most SSL algorithms use label information to improve the learning performance. However, they may also face many problems such as over-fitting. In order to improve learning machine's generation ability as much as possible, we aim to propose a semi-supervised framework that learns the predictions that do not directly depend on the label samples.

We adopt unsupervised data representation methods, such as nonnegative matrix factorization (NMF) and sparse coding (SC) as the learning machines. Without considering the difference between labeled samples and unlabeled samples, these data representation methods always have a good generation ability. Inspired by graph based method [3,4,6], which usually use an affinity graph to smooth the representation of data. We make use of label information by learning an affinity graph, which contains label information, i.e., prior information is embedded in the manifold regularization to control the smoothness of the data representation. When such an affinity graph is obtained, the corresponding graph Laplacian regularizer is incorporated into the unsupervised data representation methods.

Different from GSSL framework and manifold regularization framework [3], which always use unsupervised Gussian kernel to construct similarity graph, we utilize metric learning method [7,11,12] to produce a kernel matrix (also known as a "Gram matrix") to measure the similarity between all pairs of samples, subsequently, the similarity matrix is sparsified and reweighted to produce the final affinity graph.

Notice that in GSSL, the sparsification of affinity matrix is important since it leads to improved efficiency in the label inference stage, better accuracy and robustness to noise. In fact, spurious connections between dissimilar nodes (which tend to be in different classes) are removed and each node connects to only a few nodes. Metric learning method, which learn the similarity and dissimilarity of pairs of the labeled samples, is a good way to lead the sparsification stage more accurate (delete more "right" spurious connections). In this way, the affinity graph is more suitable with considering the label information.

In this paper, we propose a novel semi-supervised framework. Experimental results on several real benchmark datasets indicate that our semi-supervised learning framework achieves encouraging results. We compared our proposed methods with state-of-art methods, include some excellent GSSL and unsupervised methods.

The rest of the paper is organized as follows: Section 2 gives a brief review of several basic algorithms. In Section 3, we introduce our proposed framework. Experimental results are presented in Section 4. Finally, we conclude in Section 5.

## II. RELATED WORK

### A. Graph-based Semi-supervised Learning (GSSL)

GSSL techniques start with computing a similarity score between all pairs of nodes using a similarity function or kernel. Then, an algorithm is selected for finding a sparse weighted subgraph from the fully connected similarity graph. There are two typical ways to build a sparse graph: neighborhood approaches including the k-nearest and $\epsilon$ neighbors algorithms, and matching approaches such as b-matching [10].

Once a graph has been sparsified, several procedures can then be used to update the graph weights such as Binary weighting and Gaussian kernel weighting. At last, given the final affinity graph and some initial label information, GSSL algorithms diffuse the labels on the known part of the graph to the unknown nodes.

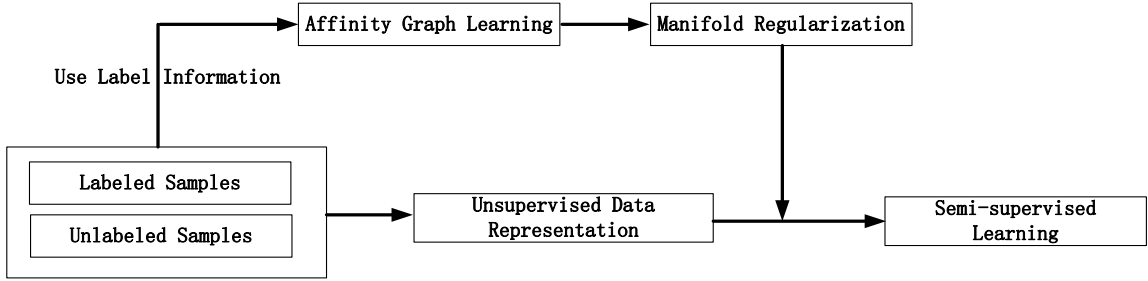

Fig. 1. The proposed framework contains two parts: 1.Affinity Graph Learning, which use both labeled and unlabeled samples to learn the local manifold structure of data. 2.Unsupervised Data Representation, which ensures the generation ability of the learning machine. Prior information is embedded in the manifold regularization to control the smoothness of the data representation.

The current best GSSL techniques include the greedy max-cut method [1], Laplacian support vector machine [8], the local and global consistency method [5] and the alternating graph transduction method [13].

*B. Metric learning*

Metric learning expects that the learned metric $M$ makes distances between similar samples small and distances between dissimilar samples large. In general, a Mahalanobis distance metric $M$ measures the squared distance between two data points $x_i$ and $x_j$:

$$d_M^2(x_i, x_j) = (x_i - x_j)^T M(x_i - x_j).$$   (1)

where $M$ is a positive semi-definite matrix and $x_i, x_j \in R^m$ is a pair of samples $(i, j)$.

The current best Metric learning techniques include Large margin nearest neighbor learning (LMNN) [11], Information theoretic metric learning (ITML) [12], and KISS metric learning [7].

III. PROPOSED FRAMEWORK

*A. Framwork*

We adopt unsupervised data representation methods as the learning machines. Traditional semi-supervised manner for unsupervised data representation usually use the Label Weight as the affinity [15]. Label Weight is constructed as follows:

$$W_{ij} = \begin{cases} 1, & y_i = y_j \\ 0, & otherwise \end{cases}.$$   (2)

In this way, the unlabeled data is totally ignored. In the proposed framework, prior information is embedded in the manifold regularization to control the smoothness of the data representation.

Given independent and identically distributed samples $X = [x_1, \ldots, x_n] \subset R^{m \times n}$ of $m$ feature dimensions and $n$ instances, and the first $l$ samples have labels. The learned weight matrix forms the following Laplacian regularizer which is used to measure the smoothness of the low dimensional representation:

$$T_0 = \frac{1}{2} \sum_{i,j=1}^{N} ||v_i - v_j||^2 W_{ij} = Tr(V^T D V) - Tr(V^T W V) = Tr(V^T L V). \tag{3}$$

where $V = [v_1; \ldots; v_n] \subset R^{n \times k}$ is the data representation of data $X$, $L \triangleq D - W$. $Tr(\cdot)$ denotes the trace of a matrix and $D$ is a diagonal matrix whose entries are column sums of $W$.

Then the loss function in the proposed framework is defined as follows:

$$T(X, R) = T_1(X, V) + \lambda Tr(V^T L V). \tag{4}$$

where $T_1(X, V)$ is the loss function of one unsupervised method, $Tr(V^T L V)$ is the Laplacian regularizer, $\lambda > 0$ is the regularization parameter to measure the smoothness of the low dimensional representation.

*B. Affinity graph learning*

We utilize KISS metric learning method [7] to produce a kernel matrix $M$ to measure the similarity between all pairs of samples. Then, the similarity scores between all pairs of nodes creates a full adjacency matrix $A \in R^{n \times n}$, where $A_{i,j} = k(x_i, x_j) = (x_i - x_j)^T M (x_i - x_j)$.

Subsequently, $k$-nearest neighbors algorithm or b-matching algorithm can be applied to build a sparse graph $P_{ij}$ ($P_{ii} = 0$). When the graph is sparsified, Gaussian kernel weighting is used to update the graph weights produce a final affinity graph. Therein, the edge weight between two connected samples $x_i$ and $x_i$ is computed as:

$$W_{ij} = P_{ij} e^{\frac{-A_{i,j}}{2\sigma^2}}. \tag{5}$$

where $\sigma$ is the kernel bandwidth parameter and we always set $\sigma = \sum_{i=1}^{n} \sum_{j=1}^{n} \sqrt{A_{i,j}^2} / n^2$.

*C. Algorithms under the proposed framework*

We adopt two powerful unsupervised data representation methods: Graph Regularized Nonnegative Matrix Factorization (GNMF) [4] and Graph Regularized Sparse Coding (GSC) [6] to display our framework.

*1)* Semi-supervised Graph Regularized Nonnegative Matrix Factorization

GNMF extend NMF by explicitly considering the manifold assumption [3] and the locally invariant idea , i.e., the nearby points are likely to have similar embeddings.

GNMF is extended to a semi-supervised method by using the proposed affinity graph. Given a nonnegative data matrix $X$, let $U = [u_1, ..., u_k] \subset R^{m \times k}$ be the basis matrix, and $V = [v_1; ...; v_n] \subset R^{n \times k}$ be the coefficient matrix. The loss function in GNMF is defined as follows:

$$T_1(X, UV^T) = T_1(X, V) + \lambda_1 Tr(V^T LV) = ||X - UV^T||^2 + \lambda_1 Tr(V^T LV).$$

$$s.t. \quad U \succcurlyeq 0, V \succcurlyeq 0. \tag{6}$$

where the regularization parameter $\lambda_1 > 0$.

The Euclidian distance based GNMF algorithms is:

$$u_{ik} \leftarrow u_{ik} \frac{(XV)_{ik}}{(UV^T V)_{ik}}.$$

$$v_{jk} \leftarrow v_{jk} \frac{(X^T U + \lambda_1 WV)_{jk}}{(VUU^T + \lambda_1 DV)_{jk}}. \tag{7}$$

*2)* Semi-supervised Graph Regularized Sparse Coding

Graph regularized Sparse Coding (GSC) learns the sparse representations that explicitly take into account the local manifold structure of the data. By incorporating the new Laplacian regularizer into the original sparse coding, GSC is extended to its semi-supervised manner. Given a data matrix $X$, let $B = [b_1, ..., b_k] \subset R^{m \times k}$ be the dictionary matrix, and $S = [s_1, ..., s_n] \subset R^{k \times n}$ be the coefficient matrix. The loss function in GSC is defined as follows:

$$T_2(X, BS) = ||X - BS||^2 + \lambda_2 Tr(SLS^T) + \lambda_3 \sum_{i=1}^{n} ||s_i||_1.$$

$$s.t. \quad ||b_i||^2 \leq c, i = 1,2, ..., k \tag{8}$$

where $\lambda_2 > 0$, $\lambda_3 > 0$ are the regularization parameter.

Following the iteratively optimization method in [6], the GSC algorithm can learn the graph regularized sparse codes S and the learning dictionary $B$ iteratively.

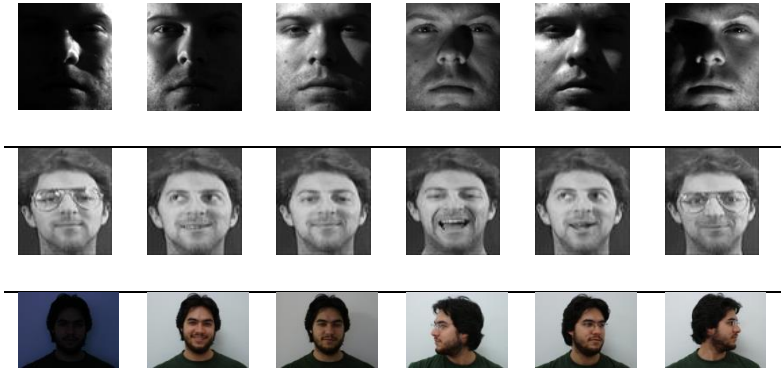

Fig. 2. Sample images in ORL, Yale and FEI database. Form the top to the bottom, each row corresponds to the Yale, ORL and FEI database, respectively.

## IV. EXPERIMENTS

Previous studies show that GNMF, GSC and GMC are very powerful for clustering, especially in the document clustering and image clustering tasks [1], [4], [6]. We investigate the clustering performance of the proposed semi-supervised framework on three real world image data sets, i.e., the Yale database[1] ,the ORL database[2], and the FEI database[3].

The ORL face database consists of 10 different images for each of 40 distinct subjects, which are taken at different times, under different lighting condition, with different facial expression and with/without glasses. The Yale database contains 165 grayscale images of 15 individuals. There are 11 images per subject, one per different facial expression or configuration. The FEI face database is taken against a white homogenous background in an upright frontal position with profile rotation of up to about 180 degrees. Scale might vary about 10%. We select 50 individuals (part 1 in FEI) for clustering experiments, and there are 14 images per subject. All images in three databases are down sampled to a size of $24 \times 32$, $32 \times 32$, $32 \times 32$ pixels with 256 gray levels per pixel, respectively. Pixel features are used for testing.

We compare the following algorithms for data clustering:

- The K-means clustering algorithm (K–means).

- Constrained Nonnegative Matrix Factorization (CNMF) [15].

- The Local and Global Consistency method (LGC) [5].

- Greedy Max–Cut (GMC) [1].

---

[1] http://homepages.dcc.ufmg.br/~william/datasets.html.
[2] http://cvc.yale.edu/projects/yalefaces/yalefaces.html.
[3] http://www.uk.research.att.com/facedatabase.html.

- Graph regularized Nonnegative Matrix Factorization （GNMF）[4] +K–means.

- Graph regularized Nonnegative Matrix Factorization using the Label Weight (2) (LGNMF) + K–means

- Graph regularized Nonnegative Matrix Factorization under the Proposed Framework (FGNMF)+ K–means.

- Graph regularized Sparse Coding (GSC) [6]+ K–means.

- Graph regularized Sparse Coding using the Label Weight (2) (LGSC) + K–means.

- Graph regularized Sparse Coding under the Proposed Framework (FGSC) + K–means.

The clustering result is evaluated by comparing the obtained label of each sample with that provided by the data set. GSSL methods such as LGC and GMC directly propagate label from labeled data to unlabeled data. For other methods, we use Accuracy Metric (AC) [14] to evaluate the clustering performance. The AC value is obtained by mapping the cluster to the corresponding predicted label, so it can be compared with the predicted accuracy in GSSL.

Table 1 Clustering Accuracy(%) on three datasets. Top three accuracy values of each column are shown as: **First**, <u>Second</u>, *Third*. Two images are selected from each subject as label information.

| Dataset | Yale | | | ORL | | | | FEI | | | |
|---|---|---|---|---|---|---|---|---|---|---|---|
| Number of Clusters | k=5 | k=10 | k=15 | k=5 | k=10 | k=20 | k=40 | k=5 | k=15 | k=30 | k=50 |
| Test Runs | 100 | 100 | 60 | 100 | 100 | 60 | 60 | 100 | 100 | 100 | 100 |
| K-means | 33.54 | 22.84 | 18.78 | 53.94 | 49.17 | 45.34 | 42.25 | 46.81 | 42.43 | 39.89 | 35.42 |
| CNMF [15] | 55.18 | 41.71 | 36.28 | 74.20 | 63.05 | 53.20 | 42.59 | 67.82 | 50.25 | 43.81 | 37.69 |
| LGC [5] | 41.81 | 30.52 | 28.32 | 75.20 | 70.08 | 65.51 | 60.80 | *83.82* | *72.09* | 64.95 | 60.54 |
| GMC [1] | 45.53 | 35.16 | 32.97 | <u>81.16</u> | *75.35* | *69.23* | *62.87* | **87.47** | <u>75.29</u> | <u>67.39</u> | <u>62.43</u> |
| GNMF [4] | 48.47 | 33.76 | 29.78 | 72.82 | 60.14 | 47.55 | 38.38 | 61.05 | 51.94 | 45.70 | 38.65 |
| LGNMF [15] | 48.30 | 35.93 | 28.75 | 67.44 | 56.22 | 44.27 | 34.37 | 60.48 | 44.79 | 37.86 | 33.81 |
| FGNMF | <u>57.76</u> | <u>44.64</u> | *38.46* | 77.32 | <u>75.73</u> | <u>71.56</u> | <u>64.28</u> | 75.01 | 67.53 | *65.35* | *60.99* |
| GSC [6] | 51.56 | 42.37 | 36.14 | *79.64* | 74.00 | 65.77 | 58.68 | 74.48 | 66.03 | 59.41 | 54.03 |
| LGSC | *57.60* | *44.14* | <u>41.03</u> | 76.38 | 71.29 | 62.80 | 57.01 | 76.48 | 60.97 | 54.48 | 49.48 |
| FGSC | **64.87** | **51.71** | **48.23** | **91.22** | **87.21** | **82.71** | **78.08** | <u>84.51</u> | **79.37** | **73.62** | **66.27** |

All algorithms expect GSSL methods obtain a new data representations of $X$. We set the dimensionality of the new space to be the same as the number of clusters. We follow the parameter settings for all algorithms to achieve their best performance, see details in [1,4,6]. For each data set, the evaluations are conducted with different numbers of clusters. For all algorithms, we randomly choose k categories from the normalized data set ($X/max(X)$), and mix the images of these k categories as the collection X for clustering. For the fixed number of clusters $k$, we randomly pick up two images from each subject as label information. Clustering performance on 10 algorithms are shown in Table 1. When labeled samples of each subject increase from two to more, clustering performance on 7 semi-supervised algorithms are shown in Table 2.

Table 2 Clustering Accuracy(%) on three datasets with fixed *k*. Top three accuracy values of each column are shown as: **First**, <u>Second</u>, *Third*. Labeled samples of each subject of different database increase from two to more.

| Number of labeled samples of each subject | ORL, k=5 | | | Yale, k=5 | | | FEI, k=5 | | | |
|---|---|---|---|---|---|---|---|---|---|---|
| | 2 | 5 | 8 | 2 | 5 | 8 | 2 | 5 | 8 | 10 |
| Test Runs | 100 | 100 | 100 | 100 | 100 | 100 | 100 | 100 | 100 | 100 |
| CNMF [15] | 74.20 | 74.74 | 75.02 | 55.18 | *53.98* | 54.76 | 67.82 | 67.47 | 68.85 | 69.65 |
| LGC [5] | <u>75.20</u> | 75.83 | 59.81 | 41.81 | 39.59 | 26.42 | *83.82* | *83.07* | 70.94 | 73.53 |
| GMC [1] | <u>81.16</u> | 81.39 | 82.07 | 45.53 | 43.02 | 44.71 | **87.47** | **86.85** | **86.64** | **88.69** |
| LGNMF [15] | <u>67.44</u> | 56.06 | 52.66 | 48.30 | 38.43 | 31.18 | 60.48 | 49.05 | 38.25 | 34.17 |
| FGNMF | *77.32* | <u>86.84</u> | <u>93.48</u> | <u>57.76</u> | <u>71.12</u> | <u>79.41</u> | 75.01 | 80.58 | *82.35* | <u>85.45</u> |
| LGSC | 76.38 | *82.74* | *85.22* | *57.60* | 62.45 | *71.52* | 76.48 | 78.11 | 76.27 | 76.28 |
| FGSC | **91.22** | **97.76** | **99.38** | **64.87** | **80.00** | **84.47** | <u>84.51</u> | <u>85.04</u> | <u>82.98</u> | *83.58* |

These experiments reveal a number of interesting points:

- The unsupervised methods under our framework outperform themselves and their original semi- supervised manner (using the Label Weigh).

- The unsupervised methods under our framework perform better than the best GSSL methods (LGC, GMC) especially when number of clusters becomes larger.

- When labeled samples of each subject increase, the unsupervised methods under our framework outperform the other semi-supervised algorithms. It seems that some algorithms like LGC and LGNMF are not sensitive to label information.

- Regardless of the data sets, the Graph Sparse Coding method under our framework always performs good. This suggests that the sparsity is important for data representation.

## V. CONCLUSIONS

We present a novel semi-supervised framework that explicitly considers the generation ability of learning machines and the prior information. By learning the affinity graph, the graph Laplacian regularizer is incorporated into the unsupervised data representation method. The experimental results on clustering have demonstrated that our proposed framework can have better discriminating power and significantly enhance the data representation performance.

## VI. ACKNOWLEDGMENTS

This paper is supported by College of Information System and Management, National University of Defense Technology and subsidized by National Natural Science Foundation of China Grant No. 61170158 and Open Project Program of the State Key Laboratory of Mathematical Engineering and Advanced Computing Grant 2013A08.